\documentclass[letterpaper]{article}
\usepackage{proceed2e}
\usepackage[margin=1in]{geometry}

\usepackage{hyperref}

\usepackage{times}

\usepackage{times}
\usepackage{latexsym}

\usepackage{url}
\usepackage{graphicx}  
\usepackage{xspace} 

\usepackage{subfigure}
\usepackage{float}
\usepackage{amsmath}
\usepackage{amsfonts}
\usepackage{amsthm}

\usepackage[flushleft]{threeparttable}

\newcommand{\kblrn}{\textsc{KBlrn}\xspace}

\title{\textsc{KBlrn}: End-to-End Learning of Knowledge Base Representations with\\ Latent, Relational, and Numerical Features}

\author{
  Alberto Garc{\'{\i}}a{-}Dur{\'{a}}n \\
  NEC Labs Europe\\
  Heidelberg, Germany \\
  \texttt{alberto.duran@neclab.eu} \\
  \And
  Mathias Niepert \\
  NEC Labs Europe\\
  Heidelberg, Germany \\
  \texttt{mathias.niepert@neclab.eu}} 

\begin{document}

\maketitle

\begin{abstract}
We present \kblrn, a framework for end-to-end learning of knowledge base representations from latent, relational, and numerical features. \kblrn integrates feature types with a novel combination of neural representation learning and probabilistic product of experts models. To the best of our knowledge, \kblrn is the first approach that learns representations of knowledge bases by integrating latent, relational, and numerical features. We show that instances of \kblrn outperform existing methods on a range of knowledge base completion tasks. We contribute a novel data set enriching commonly used knowledge base completion benchmarks with numerical features. The data sets are available under a permissive BSD-3 license\footnote{https://github.com/nle-ml/mmkb}. We also investigate the impact numerical features have on the KB completion performance of \kblrn.
\end{abstract}

\section{Introduction}

The importance of knowledge bases (KBs) for AI systems has been demonstrated in numerous application domains. KBs provide ways to organize, manage, and retrieve structured data and allow AI system to perform reasoning.  In recent years, KBs have been playing an increasingly crucial role in AI applications. Purely logical representations of knowledge bases have a long history in AI~\cite{Russell:2009}. However, they suffer from being inefficient and brittle. Inefficient because the computational complexity of reasoning is exponential in the worst case and, therefore, the time required by a reasoner highly unpredictable. Brittle because a purely logical KB requires a large set of logical rules that are handcrafted and/or mined. These problems are even more pressing in applications whose environments are changing over time.

Motivated by these shortcomings, there has been a flurry of work on combining logical and statistical approaches to build systems capable of reasoning over and learning from incomplete  structured data. Most notably, the statistical relational learning community has proposed numerous formalisms that combine logic and probability~\cite{raedt2016statistical}. These formalisms are able to address the learning problem and make the resulting AI systems more robust to missing data and missing rules. Intuitively, logical formulas act as relational features and the probability of a possible world is determined by a sufficient statistic for the values of these features. These approaches, however, are in in most cases even less efficient because logical inference is substituted with probabilistic inference.

\begin{figure}[t!]
\centering
\includegraphics[width=0.42\textwidth]{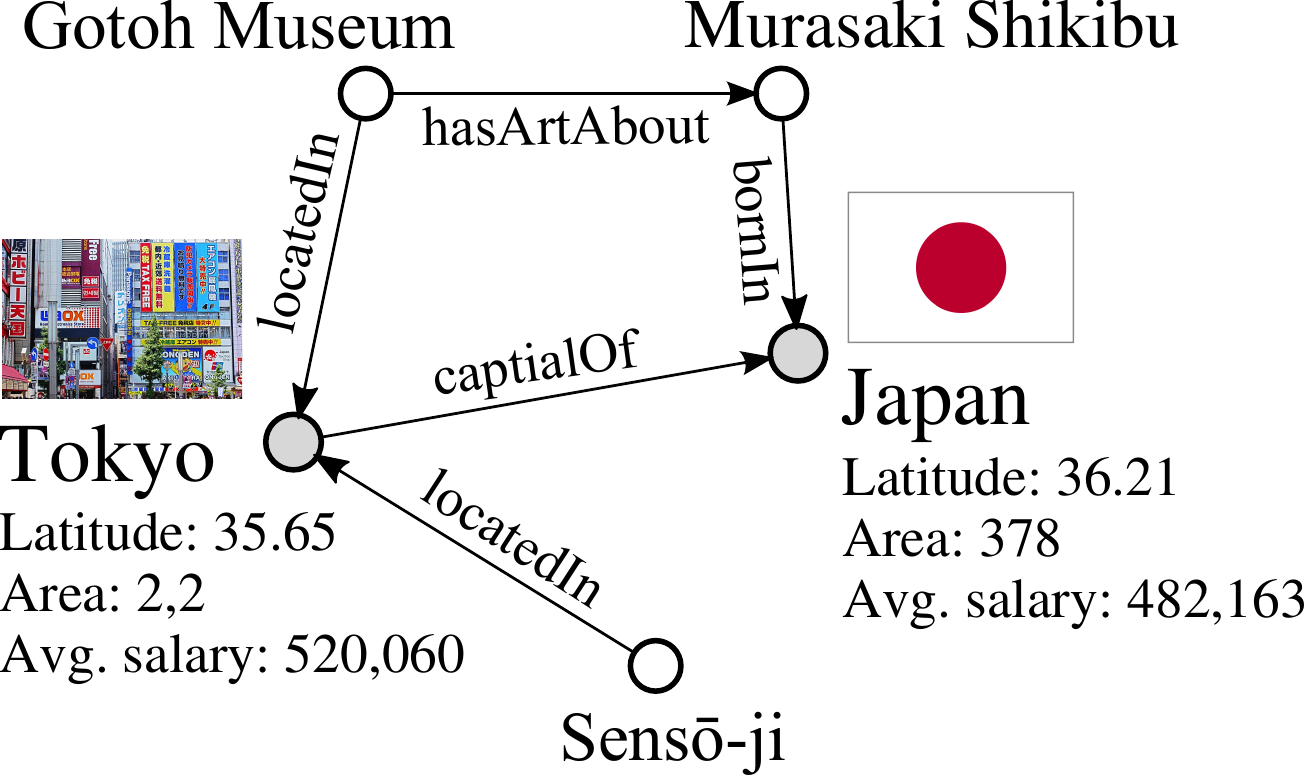}
\caption{\label{fig:KB-fragment} A small part of a knowledge base.}
\end{figure}

More recently, the research community has focused on efficient machine learning models that perform well on restricted tasks such as link prediction in KBs. Examples are knowledge base factorization and embedding approaches~\cite{bordes2013translating,nickel2011three,guu2015traversing,nickel2016review} and random-walk based ML models~\cite{lao2011random,gardner2015efficient}. The former learn latent features for the entities and relations in the knowledge base and use those to perform link prediction. The latter explore specific relational features such as path types between two entities and train a machine learning model for link prediction. 

With this work, we propose \kblrn, a novel approach to combining relational, latent (learned), and numerical features, that is, features that can take on  a large or infinite number of real values. The combination of various features types is achieved by integrating embedding-based learning with probabilistic models in two ways. First, we show that modeling numerical features with radial basis functions is beneficial and can be integrated in an end-to-end differentiable learning system. Second, we propose a probabilistic product of experts (PoE)~\cite{hinton2002training} approach to combine the feature types. Instead of training the PoE with contrastive divergence, we approximate the partition function with a negative sampling strategy. The PoE approach has the advantage of being able to train the model jointly and end-to-end.  

The paper is organized as follows. First, we discuss relational, latent, and numerical features. Second, we describe \kblrn. Third, we present empirical evidence that instances of \kblrn outperform state of the art methods for KB completion. We also investigate in detail under what conditions numerical features are beneficial.

\section{Relational, Latent, and Numerical Features}
\label{multiLearning}

We assume that the facts of a knowledge base (KB) are given as a set of triples of the form $(\mathtt{h}, \mathtt{r}, \mathtt{t})$ where $\mathtt{h}$ and $\mathtt{t}$ are the head and tail entities and $\mathtt{r}$ is a relation type. Figure~\ref{fig:KB-fragment} depicts a small fragment of a KB with relations and numerical features. KB completion is the problem of answering queries of the form $(?, \mathtt{r}, \mathtt{t})$ or $(\mathtt{h}, \mathtt{r}, ?)$. While the proposed approach can be generalized to more complex queries, we focus on completion queries for the sake of simplicity. We now discuss the three feature types used in \kblrn and motivate their utility for knowledge base completion. How exactly we extract features from a given KB is described in the experimental section.

\subsection{Relational Features}

Each relational feature is given as a logical formula which is evaluated in the KB to determine the feature's value. For instance, the formula $\exists x\ (\mathtt{A}, \mathtt{bornIn}, x) \wedge (x, \mathtt{capitalOf}, \mathtt{B})$ corresponds to a binary feature which is $1$ if there exists a path of that type from entity $\mathtt{A}$ to entity $\mathtt{B}$, and $0$ otherwise. These features are often used in relational models~\cite{toutanova2015observed,niepert2016} and random-walk based models such as PRA and SFE~\cite{lao2011random,gardner2015efficient}. 
In this work, we use relational paths of length one and two and use the rule mining approach \textsc{Amie+}~\cite{Galarraga:2015}. We detail the generation of the relational features in the experimental section. For a pair of entities $(\mathtt{h},\mathtt{t})$, we denote the feature vector computed based on a set of relational features by $\mathtt{r}_{(\mathtt{h},\mathtt{t})}$.

\begin{figure*}[t!]
\centering
\includegraphics[width=1.0\textwidth]{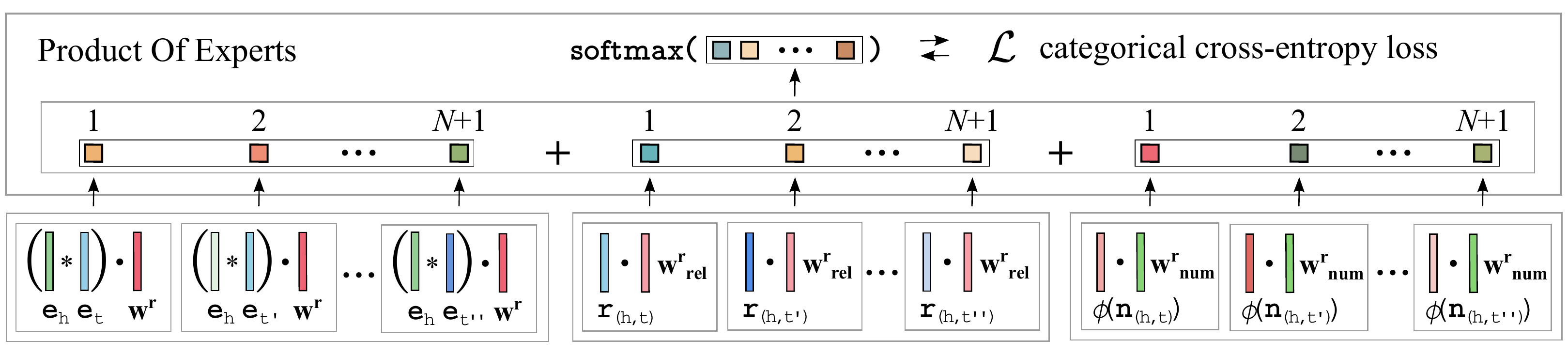}
\caption{\label{kblrn-model}An illustration of an instance of \kblrn implemented with standard deep learning framework components. For every relation type, there is a separate expert for each of the different feature types. The entities $\mathtt{t}'$ and $\mathtt{t}''$ are two of $N$ randomly sampled entities. The scores of the various submodels are added and normalized with a softmax function. A categorical cross-entropy loss is applied to the normalized scores.}
\end{figure*}

\subsection{Latent Features}

Numerous embedding methods for KBs have been proposed in recent years \cite{nickel2011three,bordes2013translating,guu2015traversing,nickel2016review}. Embedding methods provide fixed-size vector representations (embeddings) for all entities in the KB. In the simplest of cases, relations are modeled as translations in the entity embedding space~\cite{bordes2013translating}. We incorporate typical embedding learning objectives into \kblrn and write $\mathbf{e}_{\mathtt{h}}$ and $\mathbf{e}_{\mathtt{t}}$ to refer to an embedding of a head entity and a tail entity, respectively. The advantages of latent feature models are their computational efficiency and their ability to learn latent entity types suitable for downstream ML tasks without hand-crafted or mined logical rules. 

\subsection{Numerical Features}

Numerical features are entity features whose values can take on a very large or infinite number of real values. To the best of our knowledge, there does not exists a principled approach that integrates numerical features into a relational ML model for KB completion. This is surprising, considering that numerical data is available in almost all existing large-scale KBs. 
The assumption that numerical data is helpful for KB completion tasks is reasonable. For several relations types the \emph{differences} between the \textit{head} and \textit{tail} are characteristic of the relation itself. For example, while the mean difference of birth years is $0.4$ for the Freebase relation \texttt{/people/marriage/spouse}, it is $32.4$ for the relation \texttt{/person/children}. These observations motivate specifically the use of \emph{differences} of numerical feature values. 
Taking the difference has the advantage that even if a numerical feature is not distributed according to a normal distribution (e.g., birth years in a KB), the difference is often normally distributed. This is important as we need to fit simple parametric distributions to the sparse numerical data. We detail the fully automated extraction and generation of the numerical features in the experimental section.

\section{\kblrn: Learning End-To-End Joint Representations for Knowledge Bases}
\label{kblrn}

With \kblrn we aim to provide a framework for end-to-end learning of KB representations. Since we want to combine different feature types (relational, latent or learned, and numerical) we need to find a suitable method for integrating the respective submodels, one per feature type. We propose a product of experts (PoE) approach where one expert is trained for each (relation type, feature type) pair. We extend the product of experts approach in two novel ways. First, we create dependencies between the experts by sharing the parameters of the entity embedding model across relation types. By doing this, we combine a probabilistic model with a model that learns vector representations from discrete and numerical data. Second, while product of experts are commonly trained with contrastive divergence~\cite{hinton2002training}, we train it with negative sampling and a cross-entropy loss. 

In general, a PoE's probability distribution is
$$p( \mathbf{d} \mid \theta_1, ..., \theta_n) = \frac{\prod_m f_m(\mathbf{d} \mid \theta_m)}{\sum_{\mathbf{c}} \prod_m f_m(\mathbf{c} \mid \theta_{m})},$$
where $\mathbf{d}$ is a data vector in a discrete space, $\theta_m$ are the parameters of individual model $m$, $f_m( \mathbf{d} \mid \phi_m)$ is the value of $d$ under model $m$, and the $\mathbf{c}$'s index all possible vectors in the data space. The PoE model is now trained to assign high probability to observed data vectors. 

In the KB context, the data vector $\mathbf{d}$ is always a triple $\mathtt{d} = (\mathtt{h}, \mathtt{r}, \mathtt{t})$ and the objective is to learn a PoE that assigns high probability to true triples and low probabilities to triples assumed to be false. 
If $(\mathtt{h},\mathtt{r},\mathtt{t})$ holds in the KB, the pair's vector representations are used as  positive training examples. 
Let $\mathtt{d} = (\mathtt{h},\mathtt{r}, \mathtt{t})$. We can now define one individual expert $f_{(\mathtt{r},\mathtt{F})}(\mathtt{d} \mid \phi_{(\mathtt{r},\mathtt{F})})$ for each (relation type $\mathtt{r}$, feature type $\mathtt{F}$) pair. 

The specific experts we utilize here are based on simple linear models and the DistMult embedding method.
\begin{equation*}
\label{eq:KB-C}
\begin{split}
f_{(\mathtt{r},\mathtt{L})}(\mathtt{d} \mid \theta_{(\mathtt{r},\mathtt{L})}) = & \ \exp(\left(\mathbf{e}_{\mathtt{h}} * \mathbf{e}_{\mathtt{t}}\right)\cdot \mathbf{w}^{\mathtt{r}})    \\
f_{(\mathtt{r},\mathtt{R})}(\mathtt{d} \mid \theta_{(\mathtt{r},\mathtt{R})}) = & \ \exp\left(\mathbf{r}_{(\mathtt{h},\mathtt{t})} \cdot \mathbf{w}^{\mathtt{r}}_{\mathtt{rel}}\right)  \\
f_{(\mathtt{r},\mathtt{N})}(\mathtt{d} \mid \theta_{(\mathtt{r},\mathtt{N})}) = & \ \exp\left(\phi\left(\mathbf{n}_{(\mathtt{h},\mathtt{t})}\right) \cdot \mathbf{w}^{\mathtt{r}}_{\mathtt{num}} \right) \mbox{ and } \\
f_{(\mathtt{r'},\mathtt{F})}(\mathtt{d} \mid \theta_{(\mathtt{r'},\mathtt{F})}) = & \ 1 \mbox { for all } \mathtt{r'} \neq \mathtt{r} \mbox { and } \mathtt{F} \in \mathtt{\{L, R, N\}},
\end{split}
\end{equation*}
where $*$ is the element-wise product, $\cdot$ is the dot product, $\mathbf{w}^{\mathtt{r}}, \mathbf{w}^{\mathtt{r}}_{\mathtt{rel}}, \mathbf{w}^{\mathtt{r}}_{\mathtt{num}}$ are the parameter vectors for the latent, relational, and numerical features corresponding to the relation $\mathtt{r}$, and $\phi$ is the radial basis function (RBF) applied element-wise to $\mathbf{n}_{(\mathtt{h},\mathtt{t})}$. $f_{(\mathtt{r},\mathtt{L})}(\mathtt{d} \mid \theta_{(\mathtt{r},\mathtt{L})})$ is equivalent to the exponential of the \textsc{DistMult}~\cite{yang2014learning} scoring function but with \kblrn we can use any of the existing KB embedding scoring functions. 

The probability for triple $\mathtt{d} = (\mathtt{h}, \mathtt{r}, \mathtt{t})$  of the PoE model is now
$$p( \mathtt{d} \mid \theta_1, ..., \theta_n) = \frac{\prod_{\mathtt{F} \in \{\mathtt{R, L, N}\}} f_{(\mathtt{r}, \mathtt{F})}(\mathtt{d} \mid \theta_{(\mathtt{r},\mathtt{F})})}{\sum_{\mathtt{c}} \prod_{\mathtt{F} \in \{\mathtt{R, L, N}\}} f_{(\mathtt{r}, \mathtt{F})}(\mathtt{c} \mid \theta_{(\mathtt{r},\mathtt{F})}))},$$
where $\mathtt{c}$ indexes all possible triples.  


For numerical features, an activation function should fire when the difference of values is in a specific \emph{range}. For example,  we want the activation to be high when the difference of the birth years between a parent and its child is close to $32.4$ years. Commonly used activation functions such as sigmoid or tanh are not suitable here, since they saturate whenever they exceed a certain threshold. 
For each relation $\mathtt{r}$ and the $d_n$ corresponding relevant numerical features, therefore, we apply a radial basis function over the differences of values 
$\phi(\mathbf{n}_{(\mathtt{h},\mathtt{t})}) = [\phi(\mathbf{n}_{(\mathtt{h},\mathtt{t})}^{(1)}), \dots, \phi(\mathbf{n}_{(\mathtt{h},\mathtt{t})}^{(d_n)})],$
where 
$$\phi\left(\mathbf{n}_{(\mathtt{h},\mathtt{t})}^{(i)}\right) = \exp\left(\frac{-||\mathbf{n}_{(\mathtt{h},\mathtt{t})}^{(i)} - c_i||_2^2}{\sigma_i^2}\right).$$ This results in the RBF kernel being activated  whenever the difference of values is close to the expected value $c_i$. We discuss and evaluate several alternative strategies for incorporating numerical features in the experimental section. 


\subsection{Learning}

Product of experts are usually trained with contrastive divergence (CD)~\cite{hinton2002training} which relies on an approximation of the gradient of the log-likelihood using a short Markov chain started at the current seen example. The advantage of CD is that the partition function, that is, the denominator of the probability $p$, which is intractable to compute, does not need to be approximated. Due to the parameterization of the PoE we have defined here, however, it is not possible to perform CD since there is no  way to sample a hidden state given a triple $\mathtt{d}$.  Hence, instead of using CD, we approximate the partition function by performing negative sampling. 

The logarithmic loss for the given training triples $\mathbf{T}$ is defined as
$$\mathcal{L} = -\sum_{\mathtt{t} \in \mathbf{T}} \log p(\mathtt{t} \mid \theta_1, ..., \theta_n).$$
 To fit the PoE to the training triples, we follow the derivative of the log likelihood of each observed triple $\mathtt{d}\in \mathbf{T}$ under the PoE
\begin{equation*}
\begin{split}
\frac{\partial \log p( \mathtt{d} \mid \theta_1, ..., \theta_n) }{ \partial \theta_m } = & \frac{\partial \log f_m(\mathtt{d} \mid \theta_m)}{\partial \theta_m} \\
- & \frac{\partial \log \sum_{\mathtt{c}} \prod_m f_m(\mathtt{c} \mid \theta_{m})}{\partial \theta_m} 
\end{split}
\end{equation*}
Now, to approximate the intractable second term of the right hand side of the above equation, we generate for each triple $\mathtt{d} = (\mathtt{h}, \mathtt{r}, \mathtt{t})$ a set $\mathbf{E}$ consisting of $N$ triples $(\mathtt{h}, \mathtt{r}, \mathtt{t'})$  by sampling  exactly $N$ entities $\mathtt{t'}$ uniformly at random from the set of all entities. The term 
$$\frac{\partial \log \sum_{\mathtt{c}} \prod_m f_m(\mathtt{c} \mid \theta_{m})}{\partial \theta_m}$$  is then approximated by 
the term 
$$\frac{\partial \log \sum_{\mathtt{c} \in \mathbf{E}} \prod_m f_m(\mathtt{c} \mid \theta_{m})}{\partial \theta_m}.$$ Analogously for the head of the triple. This is often referred to as negative sampling.   Figure~\ref{kblrn-model} illustrates the \kblrn framework. 

\begin{table*}[t!]
\small
\centering
\begin{tabular}{|l|r|r|r|r|r|r|}
\hline
Data set & FB15k & FB15k-num & FB15k-237 & FB15k-237-num & WN18  & FB122 \\
\hline
Entities    & 14,951 & 14,951 & 14,541 & 14,541 & 40,943 & 9,738 \\
Relation types & 1,345 & 1,345 & 237 & 237 &18 & 122 \\
Training triples    & 483,142 & 483,142 & 272,115 & 272,115 &141,442 & 91,638\\
Validation triples    & 50,000 & 5,156 & 17,535 & 1,058 &5,000 & 9,595 \\
Test triples     & 59,071 & 6,012 & 20, 466  & 1,215 &5,000 & 11,243 \\
Relational features & 90,318 & 90,318 & 7,834 &  7,834 & 14 & 47\\
\hline
\end{tabular}
\caption{\label{tab:StatsKB} Statistics of the data sets.}
\end{table*}

\section{Related Work}
\label{related}

A combination of latent and relational features has been explored by Toutanova et al. \cite{toutanova2015observed,toutanova2015representing}. There, a weighted combination of two independently learned models, a latent feature model \cite{yang2014learning} and a model fed with a binary vector reflecting the presence of paths of length one between the \textit{head} and \textit{tail}, is proposed. These simple relational features aim at capturing association strengths between pairs of relationships (e.g. contains and contained$\_$by). Riedel et al. \cite{riedel2013relation} proposed a method that learns implicit associations between pairs of relations in addition to a latent feature model in the context of relation extraction. Gardner et al. \cite{gardner2014incorporating} modifies the path ranking algorithm (PRA)~\cite{lao2011random} to incorporate latent representations into models based on random walks in KBs. Gardner et al. \cite{gardner2015efficient} extracted relational features other than paths to better capture entity type information. There are a number of recent approaches that combine relational and latent representations by incorporating known logical rules into the embedding learning formulation~\cite{rocktaschel2015injecting,guo2016jointly,minervini2017adversarial}. Despite its simplicity, \kblrn{'s} combination of relational and latent representations significantly outperforms all these approaches.

There exists additional work on combining various types of KB features. Nickel et al. \cite{nickel2014reducing} proposed a modification of the well-known tensor factorization method \textsc{RESCAL}~\cite{nickel2011three}, called \textsc{ARE}, which adds a learnable matrix that weighs a set of metrics (e.g. Common Neighbors) between pairs of entities; Garcia-Duran et al. \cite{garcia2015combining} proposed a combination of latent features, aiming to take advantage of the different interaction patterns between the elements of a triple. The integration of different feature types into relational machine learning models has been previously addressed \cite{andrzejewski2011framework} \cite{liu2015trust}, but not in the context of link prediction in multi-relational graphs.

\kblrn is different to these approaches in that $(\romannumeral 1)$ we incorporate numerical features for KB completion, $(\romannumeral 2)$ we propose a unifying end-to-end learning framework that integrates arbitrary relational, latent, and numerical features.

More recent work~\cite{mmkbe:akbc17} combines numerical, visual, and textual features by learning feature type specific encoders and using the vector representations in an off-the-shelf scoring function such as \textsc{DistMult}. In contrast to this approach, \kblrn combines experts that are specialized to a specific feature type with a product of expert approach. Moreover, by taking the \emph{difference} between numerical features and explicitly modeling relational features that hold \emph{between} head and tail entities, \kblrn incorporates dependencies between modalities of the head and tail entities. These dependencies cannot be captured with a model that only includes modalities of either the head or the tail entity but not both at the same time. 

\section{Experiments}
\label{sec:exps}

We conducted experiments on six different knowledge base completion data sets. Primarily, we wanted to understand for what type of relations numerical features are helpful and what input representation of numerical features achieves the best results. An additional objective was the comparison to state of the art methods.

\subsection{Datasets}

We conducted experiments on six different data sets: FB15k, FB15k-237, FB15k-num, FB15k-237-num, WN18, and FB122.
FB15k~\cite{bordes2013translating} and Wordnet (WN)~\cite{bordes2014semantic} are knowledge base completion data sets commonly used in the literature. The FB15k data set is a representative subset of the Freebase knowledge base. WN18 represents lexical relations between word senses. The two data sets are being increasingly criticized for the frequent occurrence of reverse relations causing simple relational baselines to outperform most embedding-based methods~\cite{toutanova2015observed}. For these reasons, we also conducted experiments with FB15k-237 a variant of FB15k where reverse relations have been removed~\cite{toutanova2015observed}.   FB122 is a subset of FB15k focusing on relations pertaining to the topics of ``people", ``location", and ``sports." In previous work, a set of $47$ logical rules was created for FB122 and subsequently used in experiments for methods that take logical rules into account~\cite{guo2016jointly,minervini2017adversarial}. 

The main objective of this paper is to investigate the impact of incorporating numerical features. Hence, we created two additional data set by removing those triples from FB15k's and FB15k-237's validation and test sets where numerical features are never used for the triples' relation type. Hence, the remaining test and validation triples lead to completion queries where the numerical features under consideration are potentially used. We refer to these new data sets as FB15k-num and FB15k-237-num. A similar methodology can be followed to evaluate the performance on a different set of numerical features. 

We extracted numerical data from the 1.9 billion triple Freebase RDF dump 
 by mining triples that associate entities to literals of some numerical type. For example, the relation \texttt{/location/geocode/latitude} maps entities to their latitude. 
We performed these extractions for all entities in FB15k but only kept a numerical feature if at least 5 entities had values for it. This resulted in 116 different numerical features and 12,826 entities for which at least one of the numerical features had a value. On average each entity had 2.3 numerical features with a value. Since numerical data is not available for Wordnet, we do not perform experiments with numerical features for variants of this KB. 

Each data set contains a set of triples that are known to be true (usually referred to as positive triples). Statistics of the data sets are provided in Table~\ref{tab:StatsKB}. 
Since the identifiers for entities and relations have been changed in FB13~\cite{SocherCMN13}, we could not extract numerical features for the data set and excluded it from the experiments.

\begin{table}[t!]
\small
\centering
\begin{tabular}{|c|c|c|c|c|}
\hline
$\mathbf{n}_{(\mathtt{h},\mathtt{t})}$ & MR & MRR & Hits@1 & Hits@10 \\
\hline
sign & 231 & 29.7 & 20.1 & 50.1 \\ 
RBF & \textbf{121} & \textbf{31.4} & \textbf{21.2} &  \textbf{52.3} \\
\hline 
\end{tabular}
\caption{\label{tab:exp-nf}\textsc{KBlrn} for two possible input representations of numerical features for FB15k-237-num.}
\end{table}
\normalsize

\begin{table*}[t!]
\small
\centering
\begin{tabular}{l|cccc|cccc|}
\cline{2-9} 
 & \multicolumn{4}{c|}{FB15k} & \multicolumn{4}{c|}{FB15k-237} \\ 
\cline{2-9} 
 & MR & MRR & Hits@1 & Hits@10 & MR & MRR & Hits@1 & Hits@10  \\ 
\hline
\multicolumn{1}{|l|}{\textsc{TransE}} & 51 & 44.3 & 25.1 & 76.3 & 214 & 25.1 & 14.5 & 46.3  \\
\multicolumn{1}{|l|}{\textsc{DistMult}} & - & 65.4 & 54.6 & 82.4 & - & 19.1 & 10.6 & 37.6  \\
\multicolumn{1}{|l|}{\textsc{ComplEx}} & - & 69.2  & 59.9 & 84.0  & - & 20.1 & 11.2  & 38.8  \\
\multicolumn{1}{|l|}{\textsc{Node+LinkFeat}} & - & \textbf{82.2} & - & 87.0 & - & 23.7 & - & 36.0  \\
\multicolumn{1}{|l|}{\textsc{R-GCN+}} & - & 69.6 & 60.1 &  84.2 & - & 24.8 & 15.3 & 41.7  \\
\multicolumn{1}{|l|}{\textsc{ConvE}} & 64 & 74.5 & 67.0 &  87.3 & 330 & 30.1 & \textbf{22.0} & 45.8  \\
\hline
\hline
\multicolumn{1}{|l|}{ } & \multicolumn{8}{|c|}{without numerical features} \\
\hline
\multicolumn{1}{|l|}{\textsc{KBl}} & 69 & 77.4 &  71.2 & 87.6 & 231 & 30.1 & 21.4 &  47.5 \\
\multicolumn{1}{|l|}{\textsc{KBr}} & 628 & 78.7 & \textbf{75.6} & 84.3  & 2518 & 18.0 & 12.8 & 28.5  \\
\multicolumn{1}{|l|}{\textsc{KBlr}} & 45 & 79.0 & 74.2 & 87.3 &  231 & 30.6 & \textbf{22.0} & 48.2  \\
\hline 
\multicolumn{1}{|l|}{ } & \multicolumn{8}{|c|}{with numerical features} \\
\hline
\multicolumn{1}{|l|}{\textsc{KBln}} & 66 & 78.3 & 72.6 & \textbf{87.8} & 229 & 30.4 & \textbf{22.0} &  47.0 \\
\multicolumn{1}{|l|}{\textsc{KBrn}} & 598 & 78.7 & \textbf{75.6} & 84.2 & 3303 & 18.2 & 13.0 &  28.7 \\
\multicolumn{1}{|l|}{\textsc{KBlrn}} & \textbf{44} & 79.4 & 74.8 & 87.5 &  \textbf{209}  & \textbf{30.9} & 21.9 & \textbf{49.3}   \\
\hline 
\end{tabular}
\caption{\label{tab:exp-table1} Results (filtered setting) for \kblrn and  state of the art approaches. }
\end{table*}

\begin{table*}[t!]
\small
\centering
  \begin{threeparttable} 
\begin{tabular}{l|cccc|cccc|}
\cline{2-9} 
 & \multicolumn{4}{c|}{FB15k-num} & \multicolumn{4}{c|}{FB15k-237-num} \\ 
\cline{2-9} 
 & MR & MRR & Hits@1 & Hits@10 & MR & MRR & Hits@1 & Hits@10  \\ 
\hline
\multicolumn{1}{|l|}{\textsc{TransE}} & 25 & 34.7 & 5.5 & 79.9 & 158 & 21.8 & 10.41 & 45.6  \\
\multicolumn{1}{|l|}{\textsc{DistMult}} & 39 & 72.6 & 62.1 & 89.7 & 195 & 26.4 & 16.4 & 47.3  \\
\hline
\hline
\multicolumn{1}{|l|}{ } & \multicolumn{8}{|c|}{without numerical features} \\
\hline
\multicolumn{1}{|l|}{\textsc{KBl}} & 39 & 72.6 & 62.1 &  89.7 & 195 & 26.4 & 16.4 &  47.3 \\
\multicolumn{1}{|l|}{\textsc{KBr}} & 399 & 84.7 & \textbf{81.6} & 90.1 & 3595 & 23.6 & 17.8 & 36.1  \\
\multicolumn{1}{|l|}{\textsc{KBlr}} & 28 & 85.3 & 80.3 & 92.4 & 232 & 29.3 &  19.7 &  49.2 \\
\hline 
\multicolumn{1}{|l|}{ } & \multicolumn{8}{|c|}{with numerical features} \\
\hline
\multicolumn{1}{|l|}{\textsc{KBln}} & 32 & 73.6 & 63.0 & 90.7 & 122 & 28.6 & 17.9 & 51.6  \\
\multicolumn{1}{|l|}{\textsc{KBrn}} & 68 & 84.0 & 80.6 & 90.0 & 600 & 26.1 & 19.3 &  39.7 \\
\multicolumn{1}{|l|}{\textsc{KBlrn}} & \textbf{25} & \textbf{85.9} & 81.0 & \textbf{92.9} & \textbf{121} & \textbf{31.4} & \textbf{21.2 }&  \textbf{52.3} \\
\hline 
\end{tabular}
\caption{\label{tab:exp-table2} Results (filtered) on the data sets where the test and validation sets are comprised of those triples whose type could potentially benefit from numerical features.}
  \end{threeparttable}
\end{table*}

\begin{table*}[t!]
\small
\centering
  \begin{threeparttable} 
\begin{tabular}{l|ccccc|ccccc|}
\cline{2-11} 
 & \multicolumn{5}{c|}{WN18+rules\cite{guo2016jointly}} & \multicolumn{5}{c|}{FB122-all\cite{guo2016jointly}} \\ 
\cline{2-11} 
& MR & MRR & Hits@3 & Hits@5 & Hits@10 & MR & MRR & Hits@3 & Hits@5 & Hits@10  \\ 
\hline
\multicolumn{1}{|l|}{\textsc{TransE}} &-  & 45.3 & 79.1 & 89.1 & 93.6 & - & 48.0  & 58.9  & 64.2 & 70.2   \\
\multicolumn{1}{|l|}{\textsc{TransH}} & - & 56.0 & 80.0 & 86.1 & 90.0 & - & 46.0 & 53.7 & 59.1 & 66.0   \\
\multicolumn{1}{|l|}{\textsc{TransR}} &  -& 51.4 & 69.7 & 77.5 & 84.3 & - & 40.1 & 46.4 & 52.4 & 59.3  \\
\multicolumn{1}{|l|}{\textsc{Kale-pre}} & - & 53.2 & 86.4  & 91.9 & 94.4 & - & 52.3  & 61.7 & 66.2  &  71.8 \\
\multicolumn{1}{|l|}{\textsc{Kale-joint}} & - & 66.2 & 85.5 & 90.1 & 93.0  & - & 52.3 & 61.2 & 66.4 & 72.8   \\
\multicolumn{1}{|l|}{\textsc{ComplEx}} & - & \textbf{94.2} & \textbf{94.7} & \textbf{95.0} & \textbf{95.1} & - & 64.1 &  67.3 & 69.5 & 71.9  \\
\multicolumn{1}{|l|}{\textsc{ASR-ComplEx}} & - & \textbf{94.2} & \textbf{94.7} & \textbf{95.0} & \textbf{95.1} & - & 69.8 &  71.7 & 73.6 & 75.7  \\
\hline
\hline
\multicolumn{1}{|l|}{\textsc{KBl}} & \textbf{537} & 80.8 & 92.5 & 93.7 & 94.7 &  117 &  69.5 & \textbf{74.6} & \textbf{77.2} & \textbf{80.0}  \\
\multicolumn{1}{|l|}{\textsc{KBr}} & 7113 & 72.0 & 72.1 & 72.1 & 72.1 & 2018 & 54.7 & 54.7 & 54.7 &  54.7 \\
\multicolumn{1}{|l|}{\textsc{KBlr}} & 588 & 93.6  &  94.5  & 94.8 & \textbf{95.1} & \textbf{113} & \textbf{70.2} & 74.0 & 77.0 & 79.7  \\
\hline 
\end{tabular}
\caption{\label{tab:exp-table3} Results (filtered setting) for KB completion benchmarks where logical rules are provided.}
  \end{threeparttable}
\end{table*}
\normalsize

\subsection{General Set-up}

We evaluated the different methods by their ability to answer completion queries of the form $(\mathtt{h}, \mathtt{r}, ?)$ and $(?, \mathtt{r}, \mathtt{t})$.  For queries of the form $(\mathtt{h}, \mathtt{r}, ?)$, we replaced the tail by each of the KB's entities in turn, sorted the triples based on the scores or probabilities returned by the different methods, and computed the rank of the correct entity. We repeated the same process for the queries of type $(?, \mathtt{r}, \mathtt{t})$. We follow the filtered setting described in \cite{bordes2013translating} which removes correct triples that are different to the target triple from the ranked list. The mean of all computed ranks is the Mean Rank (lower is better) and the fraction of correct entities ranked in the top $n$ is called hits@$n$ (higher is better). We also computer the Mean Reciprocal Rank (higher is better) which is an evaluation measure for rankings that is less susceptible to outliers. 

We conduct experiments with the scoring function of \textsc{DistMult} \cite{yang2014learning} which is an application of parallel factor analysis to multi-relational graphs. For a review on parallel factor analysis we refer the reader to~\cite{harshman1994parafac}. We validated the embedding size of \kblrn from the values $\{100,200\}$ for all experiments. These values are used in most of the literature on KB embedding methods. For all other embedding methods, we report the original results from the literature or run the authors' original implementation. For FB15k and FB15k-237, the results for DistMult, Complex, and R-GCN+ are taken from~\cite{schlichtkrull2017modeling}; results for the relational baseline Node+LinkFeat are taken from~\cite{toutanova2015observed}; results for ConvE are taken from \cite{dettmers2017convolutional} and results for TransE were obtained by running the authors' implementation. 
For WN18-rules and FB122-all, the results for TransE, TransH, TransR, and KALE are taken from~\cite{guo2016jointly}, and results for ComplEx and ASR-ComplEx are taken from~\cite{minervini2017adversarial}. All methods were tuned for each of the respective data sets. 

For \kblrn we used \textsc{Adam} \cite{kingma2014adam} for parameter learning in a mini-batch setting with a learning rate of $0.001$, the categorical cross-entropy as loss function and the number of epochs was set to 100. We validated every 5 epochs and stopped learning whenever the MRR (Mean Reciprocal Rank) values on the validation set decreased. The batch size was set to $512$ and the number $N$ of negative samples to 500 for all experiments. 
We use the abbreviations KB\textit{suffix} to refer to the different instances of \kblrn{.} \textit{suffix} is a combination of the letters \textsc{l} (\textsc{l}atent), \textsc{r} (\textsc{r}elational) and \textsc{n} (\textsc{n}umerical) to indicate the inclusion of each of the three feature types. 

\subsection{Automated Generation of Relational and Numerical Features}

For the data sets FB15k, FB15k-237, and their numerical versions, we used all relational paths of length one and two found in the training data as relational features. These correspond to the formula types $(\mathtt{h}, \mathtt{r}, \mathtt{t})$ (1-hop) and $\exists x\ (\mathtt{h}, \mathtt{r}_1, x) \wedge (x, \mathtt{r}_2, \mathtt{t})$ (2-hops). We computed these relational paths with \textsc{Amie+}~\cite{Galarraga:2015}, a highly efficient system for mining logical rules from knowledge bases. We used the standard settings of \textsc{Amie+} with the exception that the minimal head support was set to $1$.  With these settings, \textsc{Amie+} returns horn rules of the form $\mathtt{body}\Rightarrow (x, \mathtt{r}, y)$ that are present for at least 1\% of the triples of the form $(x, \mathtt{r}, y)$. For each relation $\mathtt{r}$, we used the body of those rules where $\mathtt{r}$ occurs in the head as $\mathtt{r}$'s relational path features. For instance, given a rule such as $(x, \mathtt{r_1}, z), (z, \mathtt{r_2}, y) \Rightarrow (x,\mathtt{r}, y)$, we introduce the relational feature $\exists x\ (\mathtt{h}, \mathtt{r}_1, x) \wedge (x, \mathtt{r}_2, \mathtt{t})$ for the relation $\mathtt{r}$. Table~\ref{tab:amie-features} lists a sample of relational features that contributed positively to the performance of \kblrn for specific relation types.
For the data sets WN18 and FB122, we used the set of logical formulas previously used in the literature~\cite{guo2016jointly}. Using the same set of relational features allows us to compare \kblrn with existing approaches that incorporate logical rules into the embedding learning objective~\cite{guo2016jointly,minervini2017adversarial}.

For each relation $\mathtt{r}$ we only included a numerical feature if, in at least $\tau=90\%$ of training triples, both the head and the tail had a value for it. This increases the likelihood that the feature is usable during test time. 
For $\tau=90\%$ there were $105$ relations in FB15k for which at least one numerical feature was included during learning, and $33$ relations in FB15k-237. With the exception of the RBF parameters, all network weights are initialized following~\cite{glorot2010understanding}. 
The parameters of \kblrn's RBF kernels are initialized and fixed to $c_i = \frac{1}{|\mathbf{T}|} \sum_{(\mathtt{h},\mathtt{r},\mathtt{t})\in\mathbf{T}} \mathbf{n}_{(\mathtt{h},\mathtt{t})}^{(i)}$, where $\mathbf{T}$ is the set of training triples $(\mathtt{h},\mathtt{r},\mathtt{t})$ for the relation $\mathtt{r}$ for which both $\mathbf{n}_{\mathtt{h}}^{(i)}$ and $\mathbf{n}_{\mathtt{t}}^{(i)}$ have a value; and $\sigma_i = \sqrt{\frac{1}{|\mathbf{T}|} \sum_{(\mathtt{h},\mathtt{r},\mathtt{t})\in\mathbf{T}}(\mathbf{n}_{(\mathtt{h},\mathtt{t})}^{(i)} - c_i)^2}$.

\begin{table}[t!]
\small
\centering
\begin{tabular}{l|c|c|c|c|}
\cline{2-5} 
 & \multicolumn{2}{|c|}{\textsc{KBlr}} & \multicolumn{2}{|c|}{\textsc{KBlrn}} \\
\hline
\multicolumn{1}{|l|}{Relation} & MRR & H@10 & MRR & H@10\\ 
\hline
\multicolumn{1}{|l|}{\texttt{capital}$\_$\texttt{of}} &  5.7 & 13.6 & \textbf{14.6} & \textbf{18.2} \\
\hline
\multicolumn{1}{|l|}{\texttt{spouse}$\_$\texttt{of}} &  4.4 & \textbf{0.0} & \textbf{7.9} & \textbf{0.0} \\
\hline 
\multicolumn{1}{|l|}{\texttt{influenced}$\_$\texttt{by}} & 7.3 & 20.9 & \textbf{9.9} & \textbf{26.8} \\
\hline 
\end{tabular}
\caption{\label{tab:exp-perrel} MRR and hits@10 results (filtered) for \kblrn with and without numerical features in FB15k-237. Results improve for relations where the difference of the relevant features is approximately normal (see Figure~\ref{vis-diff}). }
\end{table}

\begin{table*}[t!]
\small
\centering
\begin{tabular}{|c|c|}
\hline
Relational Feature & Triple $(\mathtt{h}, \mathtt{r}, \mathtt{t})$\\
\hline
$\exists \texttt{x}$ (\texttt{h}, \texttt{containedby}, \texttt{x}) $\wedge$ (\texttt{t}, \texttt{locations$\_$in$\_$this$\_$time$\_$zone}, \texttt{x})   & $(\texttt{h},\texttt{time$\_$zone}, \texttt{t})$\\
\hline
$\exists \texttt{x}$ (\texttt{t}, \texttt{prequel}, \texttt{x}) $\wedge$ (\texttt{x}, \texttt{character}, \texttt{h})   & $(\texttt{h}, \texttt{character$\_$in$\_$film}, \texttt{t})$\\
\hline
$\exists \texttt{x}$ (\texttt{x}, \texttt{cause$\_$of$\_$death}, \texttt{h}) $\wedge$ (\texttt{x}, \texttt{cause$\_$of$\_$death}, \texttt{t})   & 
$(\texttt{h}, \texttt{includes$\_$causes$\_$of$\_$death}, \texttt{t})$\\
\hline
\end{tabular}
\caption{\label{tab:amie-features} Relational features found by \textsc{Amie+} that positively contributed to the performance of \textsc{KBLrn} for the particular relation type holding between a head entity $\mathtt{h}$ and tail entity $\mathtt{t}$.}
\end{table*}

\subsection{Representations of Numerical Features}

We experimented with different strategies for incorporating raw numerical features. For the difference of feature values the simplest method is the application of the sign function. For a numerical attribute $i$, the activation  is either $1$ or $-1$ depending on whether the difference $\mathbf{n}_{\mathtt{h}}^{(i)} - \mathbf{n}_{\mathtt{t}}^{(i)}$ is positive or negative.  
For a more nuanced representation of differences of numerical features, a layer of RBF kernels is a suitable choice since the activation is here highest in a particular range of input values. 
The RBF kernel might not be appropriate, however, in cases where the underlying distribution is not normal.


To evaluate different input representations, we conducted experiments with \textsc{KBlrn} on the FB15k-237-num data set. Table \ref{tab:exp-nf} depicts the KB completion performance of two representation strategies for the difference of head and tail values. Each row corresponds to one evaluated strategy. ``sign" stands for applying the sign function to the difference of numerical feature values. RBF stands for using an RBF kernel layer for the differences of numerical feature values. All results are for the FB15k-237-num test triples.

The RBF kernels outperform the sign functions significantly. This indicates that the difference of feature values is often distributed normally and that having a region of activation is beneficial.  
Given these results, we use the RBF input layer for $\mathbf{n}_{(\mathtt{h},\mathtt{t})}$ for the remainder of the experiments.

\begin{figure}[t!]
\centering
\subfigure{\includegraphics[width=0.16\textwidth]{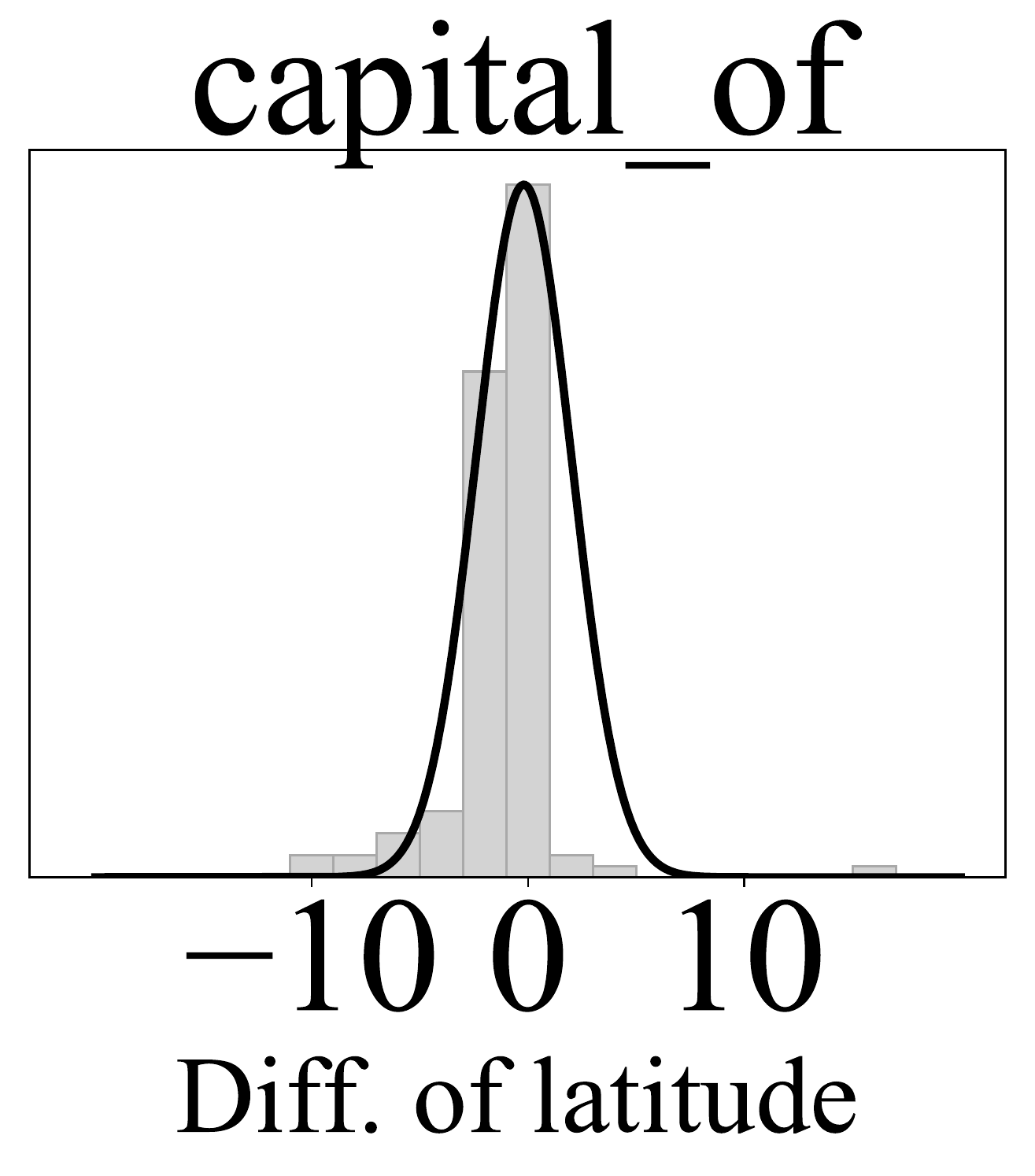}}
\hspace{-2mm}
\subfigure{\includegraphics[width=0.16\textwidth]{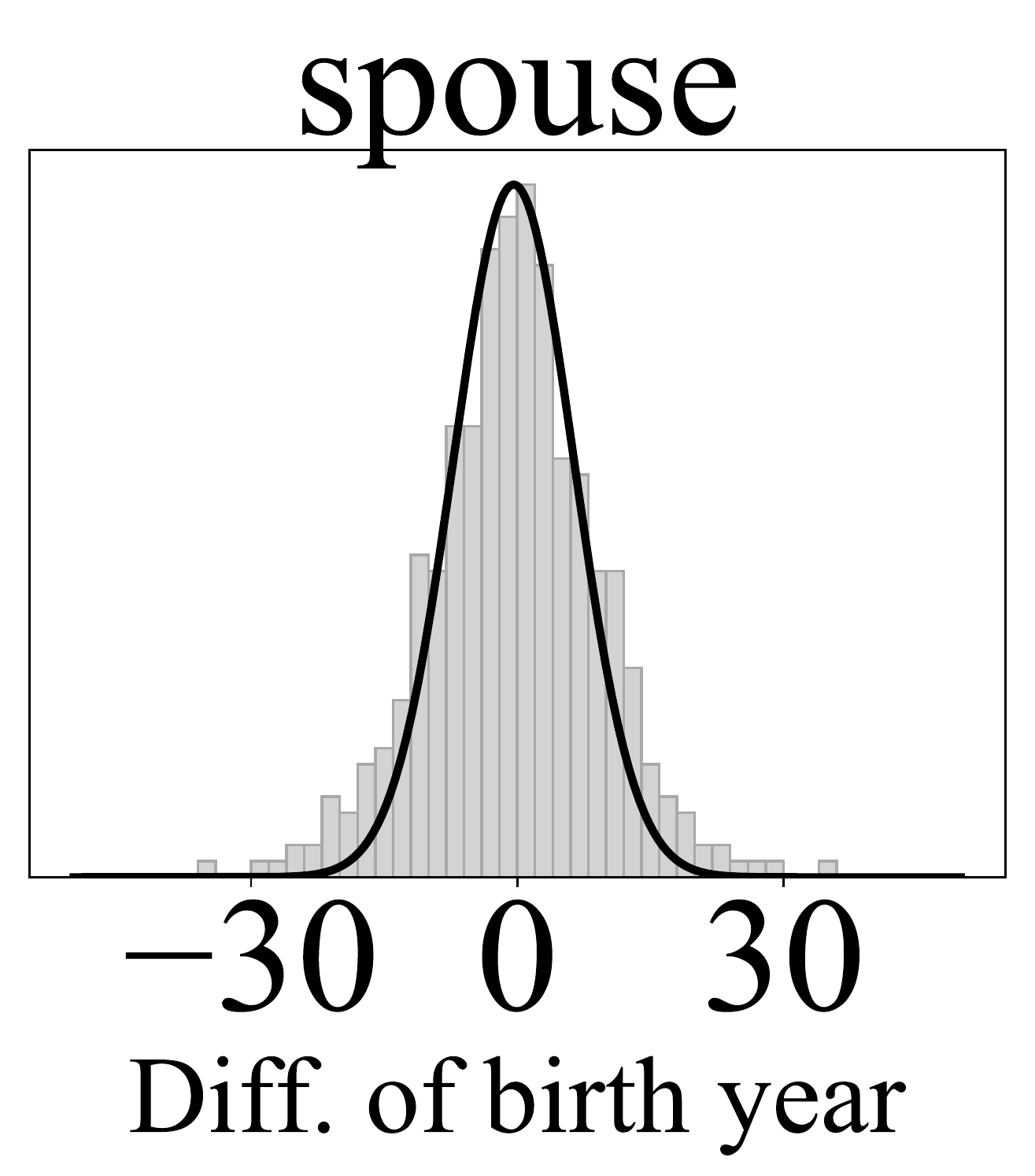}}
\hspace{-2mm}
\subfigure{\includegraphics[width=0.16\textwidth]{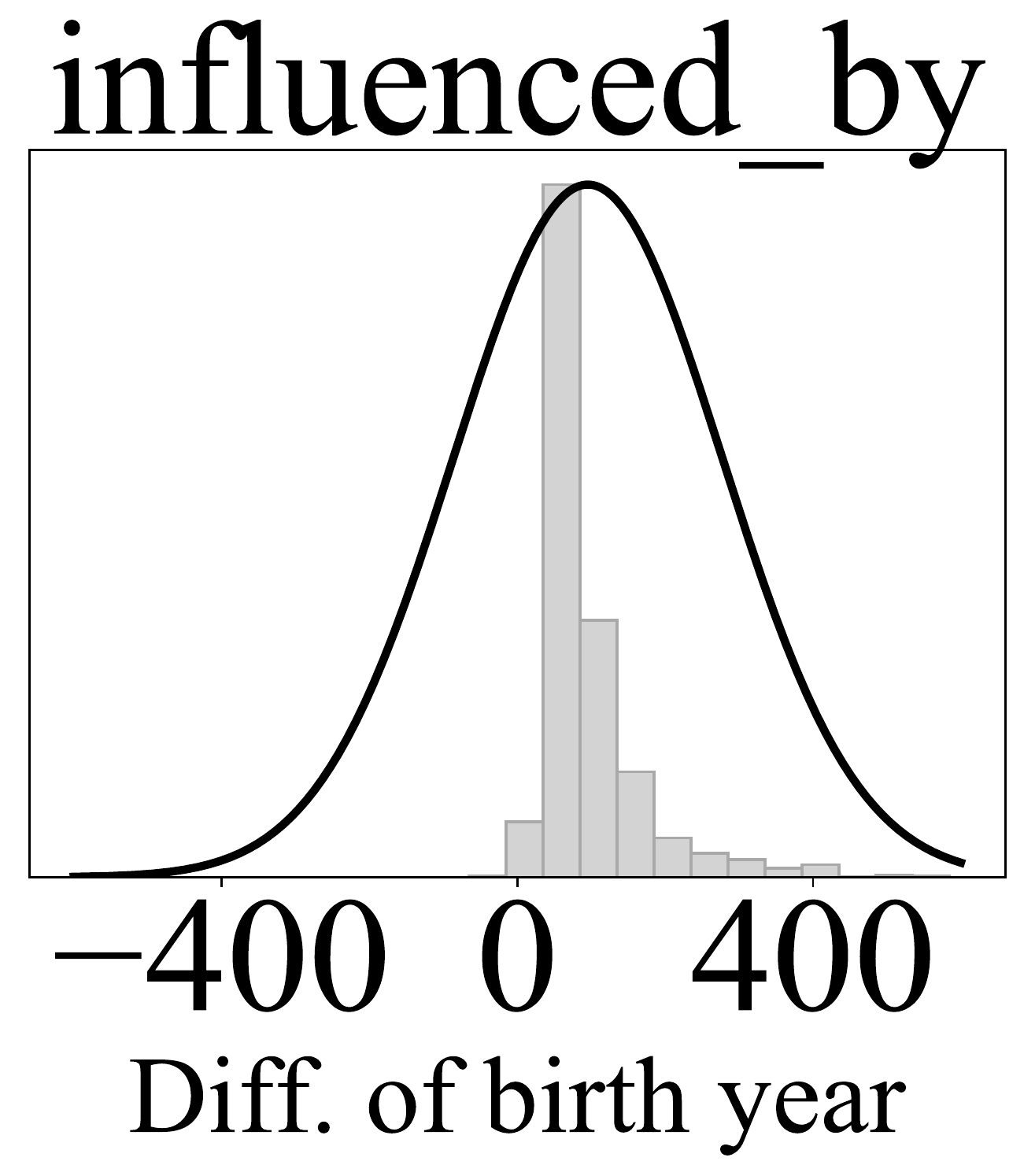}}
\caption{\label{vis-diff} Histograms and fitted RBFs for three representative relations and numerical features.}
\end{figure}

\subsection{Comparison to State of the Art}

For the standard benchmark data sets FB15k and FB15k-237, we compare \kblrn with \textsc{TransE}, \textsc{DistMult}, \textsc{ComplEx}~\cite{trouillon2016complex}, \textsc{R-GCN+}~\cite{schlichtkrull2017modeling}, and ConvE~\cite{dettmers2017convolutional}.

Table \ref{tab:exp-table1} lists the KB completion results. \kblrn is competitive with state of the art knowledge base completion methods on FB15k and significantly outperforms all other methods on the more challenging FB15k-237 data set. Since the fraction of triples that can potentially benefit from numerical features is very small for FB15k, the inclusion of numerical features  is only slightly beneficial. For FB15k-237, however, the numerical features significantly improve the results.  

For the numerical versions of FB15k and FB15k-237, we compared \kblrn to TransE and DistMult. Table \ref{tab:exp-table2} lists the results for the KB completion task on these data sets. \kblrn significantly outperforms the KB embedding approaches. The positive impact of including the numerical features is significant. 

For the data sets WN18-rules and FB122-all we compared \kblrn to KB embedding methods TransE, TransR~\cite{lin2015learning}, TransH~\cite{wang2014knowledge}, and ComplEx~\cite{trouillon2016complex} as well as state of the are approaches for incorporating logical rules into the learning process. The experimental set-up is consistent with that of previous work. Table \ref{tab:exp-table3} lists the results for the KB completion task on these data sets. \kblrn combining relational and latent representations significantly outperforms existing approaches on FB122 with exactly the same set of rules. This provides evidence that \kblrn{'s} strategy to combine latent and relational features is effective despite its simplicity relative to existing approaches. For WN18+rules, \kblrn is competitive with ComplEx, the best performing method on this data set. In addition, \kblrn{'s} performance improves significantly when relational and latent representations are combined. In contrast, \textsc{ASR-ComplEx} is not able to improve the results of ComplEx, its underlying  latent representation.

\begin{table}[t!]
\small
\begin{center}
\begin{tabular}{l|c|c|c|c|}
\cline{2-5} 
 & \multicolumn{2}{c|}{MR}  & \multicolumn{2}{c|}{MRR} \\
\cline{2-5}
 & \textsc{One}  & \textsc{Many} & \textsc{One}  & \textsc{Many} \\
\hline
\multicolumn{1}{|l|}{\textsc{KBlr}} & \textbf{42} & 201 & \textbf{74.9} & 21.0 \\
\hline
\hline
\multicolumn{1}{|l|}{\textsc{KBlrn}} & 60 & \textbf{135} & 74.2 & \textbf{22.8}\\
\hline
\end{tabular}
\end{center}
\caption{\label{tab:cardinality} MR and MRR results (filtered) on FB15k-237-num based on the cardinality of the test queries.}
\end{table}

\begin{table}[t!]
\small
\centering
\begin{tabular}{l|c|c|c|}
\cline{2-4} 
 & AUC-PR  & MR & MRR \\ 
\hline
\multicolumn{1}{|l|}{\textsc{TransE}} & 0.837 & 231 & 26.5 \\
\hline
\multicolumn{1}{|l|}{\textsc{KBlr}}  & 0.913  & 94  & 66.8 \\
\hline
\hline
\multicolumn{1}{|l|}{\textsc{KBlrn}} & \textbf{0.958} & \textbf{43} & \textbf{70.8} \\
\hline 
\end{tabular}
\caption{\label{tab:exp-fiber} AUC-PR, MR, and MRR results for the completion query (\texttt{USA}, \texttt{/location/contains}, ?).}
\end{table}
 
\subsection{The Impact of Numerical Features} 
 
The integration of numerical features improves \kblrn's performance significantly. We performed several additional experiments so as to gain a deeper understanding of the impact numerical features have. 

Table \ref{tab:exp-perrel} compares \kblrn's performance with and without integrating numerical features on three relations. The performance of the model with numerical features is clearly superior for all three relationships (\texttt{capital}$\_$\texttt{of}, \texttt{spouse} and \texttt{influenced}$\_$\texttt{by}). Figure \ref{vis-diff} shows the normalized histograms for the values $\mathbf{n}_{(\mathtt{h},\mathtt{t})}$ for these relations. We observe the differences of feature values are approximately normal. 

Following previous work~\cite{bordes2013translating}, we have classified each test query of FB15k-237-num as either \textsc{One} or \textsc{Many}, depending on the whether one or many entities can complete that query. For the queries labeled \textsc{One} the model without numerical features shows a slightly worse performance with respect to the model that makes use of them, whereas for the queries labeled \textsc{Many}, \textsc{KBlrn} significantly outperforms \textsc{KBlr} in both MR and MRR. 

A well-known finding is the lack of completeness of FB15k and FB15k-237. This results in numerous cases where the correct entity for a  completion query is not contained in the ground truth (neither in the training, nor test, nor validation data set). This is especially problematic for queries where a large number of entities are correct completions. 
To investigate the actual benefits of the numerical features we carried out the following experiment: We manually determined all correct completions for the query (\texttt{USA}, \texttt{/location/contains}, ?). We ended up with $1619$ entities that correctly complete the query. FB15k-237 contains only $954$ of these correct completions. With a complete ground truth, we can now use the precision-recall area under the curve (PR-AUC) metric to evaluate KB completion methods \cite{nickel2011three,nickel2014reducing,garcia2015combining}. A high PR-AUC represents both high recall and high precision.  Table \ref{tab:exp-fiber} lists the results for the different methods. \kblrn with numerical features consistently and significantly outperformed all other approaches. 


\section{Conclusion}

We introduced \kblrn, a class of machine learning models that, to the best of our knowledge, is the first proposal aiming at integrating relational, latent, and continuous features of head and tail entities in KBs into a single end-to-end differentiable framework. \kblrn outperforms state of the art KB completion methods on a range of data sets. We show that the inclusion of numerical features is beneficial for KB completion tasks.

Future work will primarily study instances of experts that can combine different numerical features such as life expectancy and latitude. Furthermore, combinations of multiple different embedding methods, which was shown to be beneficial in recent work~\cite{wang2017multi}, is also possible with our PoE approach.

\bibliographystyle{abbrv}
\bibliography{uai}

\end{document}